\documentclass{bmvc2k}

\usepackage{csquotes} %
\usepackage[capitalise]{cleveref}
\usepackage{xcolor}
\usepackage{comment}
\usepackage{multirow}
\usepackage{amsmath}
\usepackage{amssymb}
\usepackage{array}
\usepackage{wrapfig}
\usepackage{etoolbox}

\newtoggle{arxiv}

\definecolor{darkgreen}{rgb}{0, 0.4, 0}
\definecolor{brightred}{rgb}{1, 0, 0}

\newcommand{\psp}{pSp}

\newcommand{\wplus}{\mathcal{W}\textsuperscript{+}}
\newcommand{\wspace}{\mathcal{W}}
\newcommand{\zspace}{\mathcal{Z}}

\DeclareMathOperator{\norm}{norm}

\newcommand{\generator}{G}
\newcommand{\critic}{C}
\newcommand{\seedinitial}{i}
\newcommand{\seedtemporal}{s}
\newcommand{\sfrom}{\sim}
\newcommand{\gauss}{{\mathcal{N}(0,1)}}
\newcommand{\reals}{\mathbb{R}}

\newcommand{\hallucinator}{H}
\newcommand{\producer}{P}
\newcommand{\translator}{T}
\newcommand{\latent}{l}

\newcommand{\extractor}{E}

\newcommand{\loss}{\mathcal{L}}
\newcommand{\wgan}{\text{WGAN}}
\newcommand{\gap}{\text{GAP}}
\newcommand{\gp}{\text{GP}}

\newcommand{\plusminus}{$\pm$}

\toggletrue{arxiv}

\title{StyleVideoGAN: A Temporal Generative Model using a Pretrained StyleGAN}

\addauthor{Gereon Fox}{gfox@mpi-inf.mpg.de}{1}
\addauthor{Ayush Tewari}{atewari@mpi-inf.mpg.de}{1}
\addauthor{Mohamed Elgharib }{elgharib@mpi-inf.mpg.de}{1}
\addauthor{Christian Theobalt }{theobalt@mpi-inf.mpg.de}{1}

\addinstitution{
 Max Planck Institute for Informatics\\
 Saarland Informatics Campus\\
 Saarbr\"ucken, Germany
}

\runninghead{Fox \etal}{StyleVideoGAN}

\def\etal{\emph{et al}\bmvaOneDot}

\begin{document}

\maketitle

\begin{abstract}
Generative adversarial models (GANs) continue to produce advances in terms of the visual quality of \emph{still} images, as well as the learning of temporal correlations. However, few works manage to combine these two interesting capabilities for the synthesis of video content: Most methods require an extensive training dataset to learn temporal correlations %
, while being rather limited in the resolution and visual quality of their output.
We present a novel approach to the video synthesis problem that helps to greatly improve visual quality and drastically reduce the amount of training data and resources necessary for generating videos. Our formulation separates the \emph{spatial} domain, in which individual frames are synthesized, from the \emph{temporal} domain, in which motion is generated.
For the spatial domain we use a pre-trained StyleGAN network, the latent space of which allows control over the appearance of the objects it was trained for. The expressive power of this model allows us to embed our training videos in the StyleGAN latent space.
Our temporal architecture is then trained not on sequences of RGB frames, but on sequences of StyleGAN latent codes. The advantageous properties of the StyleGAN space simplify the discovery of temporal correlations.
We demonstrate that it suffices to train our temporal architecture on only 10 minutes of footage of  1 subject for about 6 hours. After training, our model can not only generate new portrait videos for the training subject, but also for \emph{any} random subject which can be embedded in the StyleGAN space.

\end{abstract}

\section{Introduction}
\label{sec:intro}
Recent advances of generative adversarial networks (GANs), notably StyleGAN \cite{Karras18,Karras2019,Karras2020}, show impressive capabilities in learning manifolds of photorealistic images at high resolution (up to $1024^2$). This is especially true for images of human faces.
However, these improvements are only starting to carry over to the domain of \emph{videos}:
While existing methods for videos show promising results in modeling content and motion \cite{Tulyakov18,Saito20,Weissenborn20,munoz2020,tian2021}, they usually are subject to at least a subset of the following limitations: small spatial resolution ( $\leq 128^2$); spatial artifacts ; constrained motion ; necessity of large amounts of training data ;  large computational cost for training (memory, time; see \cref{fig:comparisons} and \cref{sec:comparisons}) .

To address these problems, we present a novel approach to unconditional video generation: Our goal is to learn a generator for nearly photorealistic, high-resolution videos (up to $1024\times 1024$), by training on a video dataset that  contains no more than 10 minutes of video footage.
In addition, we limit the training footage to depict only a single subject.
However, we want the trained model can generate motion not only for the training subject, but for \emph{many} different \emph{random} subjects.
As the proving ground for our method we chose the generation of portrait videos, because (1) portraits are an attractive target for animation, and (2) high-quality training data and StyleGAN models for this domain are readily available.
To demonstrate that our method is also applicable to other domains very different from portrait, we also show how it can be applied to the generation of complex hand motion.
\begin{figure}
	\begin{center}
    		\includegraphics[width=1.0\linewidth]{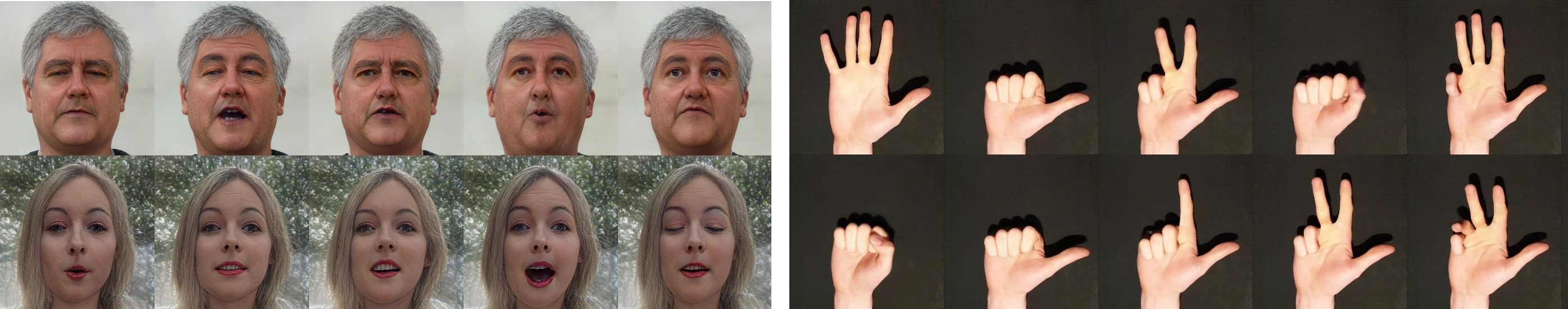}
    		\caption{ Face and hand videos generated by our method}
    		\label{fig:teaser}
	\end{center}
\end{figure}

Our key idea for addressing this task is to embed our training video set into the $\wplus{}$ latent space of a pretrained StyleGAN model \cite{richardson2021}~.
Embedding videos in $\wplus{}$ turns them from sequences of RGB frames into sequences of $\wplus{}$ vectors. While an RGB frame has
$1024 \cdot 1024 \cdot 3 = 3145728$ dimensions, a $\wplus{}$ code only has $18 \cdot 512 = 9216$ dimensions. This means that the embedding allows our generative model to be trained in a much lower-dimensional space, which simplifies the discovery of temporal correlations. Also, this transformation completely eliminates the necessity to actually render any video frames at training time, which greatly reduces the amount of memory and the time required to train our model. Our model is supervised completely in $\wplus{}$ space, unlike any other existing approach.

There is one more major advantage of the embedding approach, that allows us to make our model generate motion for a great multitude of subjects, even though it has seen only one subject at training time: The linear separability properties of the $\wplus{}$ space \cite{Tewari20Stylerig,shen2020,Nitzan20} allow us to analyze the shape of the $\wplus{}$  embedding of the training footage.
Using such an analysis, we present an \enquote{\textbf{offset trick}} which allows us to transfer the motion of a generated video to a different subject.

The ideas described so far already allow us to generate very high resolution videos with a minimal demand of training data and computational resources, by training a recurrent Wasserstein GAN \cite{arjovsky2017} on temporal volumes of 25 time steps, more than what most previous methods can afford. However, in order to have our model generate videos of longer duration, our generator needs to continue its output sequence beyond 25 time steps at test time. This can be achieved by making the generator a recurrent neural network (RNN). Previous work \cite{tian2021} has pointed out, though, that just using an RNN is not sufficient. We validate this observation by demonstrating that a vanilla RNN may tend to \enquote{loop}, i.e., repeat the same motion over and over.
We address this problem with a novel \enquote{\textbf{gradient angle penalty}} term which successfully prevents looping.
In summary, we make the following  contributions:
\begin{itemize}

	\item We present a novel approach for unconditional video generation that is supervised in the latent space of a pretrained image generator, without having to render video frames at training time, leading to large savings in computational resources.

	\item We are the first to demonstrate how the properties of StyleGAN's $\wplus{}$ space can greatly reduce the amount of training data necessary to train a video generative model.

	\item We present a novel \enquote{gradient angle penalty} loss that helps in the generation of videos that are longer than the temporal window seen at training time.

	\item We demonstrate that our approach is applicable even to domains as complicated as hand motion, where there are more challenging articulations and self-occlusions.

\end{itemize}

\section{Related Work}
\label{sec:related}

\subsection{Generative Models for Videos}

Several methods have been proposed for learning a generative model of videos~\cite{vondrick16,Saito17,Denton17,Tulyakov18,Acharya18,Yushchenko19,Clark19,Saito20,Kahembwe20,Weissenborn20,munoz2020,Ye20,menapace2021playable,Chai21,tian2021,Hong21ArrowGAN,wang2021}.
While such approaches show interesting results, they are limited to low spatial resolutions such as 128x128~\cite{Tulyakov18,munoz2020,Ye20,wang2021}, 256x256~\cite{Acharya18,Clark19,Saito20,Hong21ArrowGAN} or 512x512~\cite{Kahembwe20}.
An exception is the work of Tian~\etal~\cite{tian2021}, which can generate videos at 1024x1024.
Furthermore, most approaches struggle with generating realistic videos of long durations.
We now discuss these existing approaches in more detail:

Saito~\etal~\cite{Saito17} presented an approach for video synthesis using Wasserstein GAN losses and a novel parameter clipping method.
The network architecture, like ours, uses a shared image generator for each frame.
However, the image generator is not pretrained, and the loss functions are defined on the final video space, and not the latent space of the image generator.
Saito~\etal demonstrate results on videos upto 128x128 resolution.
Tulyakov~\etal~\cite{Tulyakov18} decompose the generation of videos into a content part and a motion part. Their \enquote{MoCoGAN} is trained in an unsupervised manner using motion and content discriminators.
Yushchenko~\etal~\cite{Yushchenko19} formulated video generation by means of Markov Decision Processes and extended into the framework of Tulyakov~\etal~\cite{Tulyakov18}.
Acharya~\etal~\cite{Acharya18} learned to progressively grow the generative model starting from low-resolution and short duration, which enabled the synthesis of videos at  256x256 resolution for the first time. %
Clark~\etal~\cite{Clark19} divided their discriminator into a spatial component and a temporal component, which also allows the generation of videos at resolutions up to 256x256 and duration up to 48 frames.
More recently, Saito~\etal~\cite{Saito20} proposed a memory-efficient approach for training that scales linearly with resolution.
Instead of directly training on the full temporal window, it uses a stack of sub-generators that are trained on different temporal resolutions.
Earlier sub-generators process high frame-rate input with low resolution information while the later sub-generators process low frame-rate input with high resolution information.

Weissenborn~\etal~\cite{Weissenborn20} proposed an autoregressive video generation model that generalizes the Transformer architecture~\cite{Vaswani17} using a three-dimensional self-attention mechanism. To reduce computational complexity, images are produced as sequences of smaller, sub-scaled image slices, akin to~\cite{Menick18}.
Munoz~\etal~\cite{munoz2020} modelled frames as points in a latent space, without any 3D convolutions. Their generator consists of a sequence generator and an image generator. Adversarial losses contain a 2D discriminator and a 3D discriminator.

Very recently, and most related to our work, Tian~\etal~\cite{tian2021} formulated video generation as the problem of finding a suitable trajectory through the latent space of a pretrained and fixed image generator~\cite{Karras2020,Brock19}, for example StyleGAN.
Despite this commonality and their ability to produce output at resolution $1024\times 1024$ as a result, there is a number of important differences between their approach and ours:
\begin{itemize}
\item Their discriminator supervises the generator in the image domain, which is a much higher-dimensional and more redundant domain than $\wplus{}$.
\item Their design inherently relies on the image generator being available for forward and backward passes at training time, which increases the required amounts of GPU memory and computation time immensely, compared to our approach.
\item Their method requires a diverse training set. When trained on a single video (like our method is), their results show very limited motion, as we demonstrate in \cref{sec:results}

\end{itemize}

\section{Method}
\label{sec:method}

\begin{wrapfigure}{R}{0.5\textwidth}
    \centering
    \includegraphics[width=0.5\textwidth]{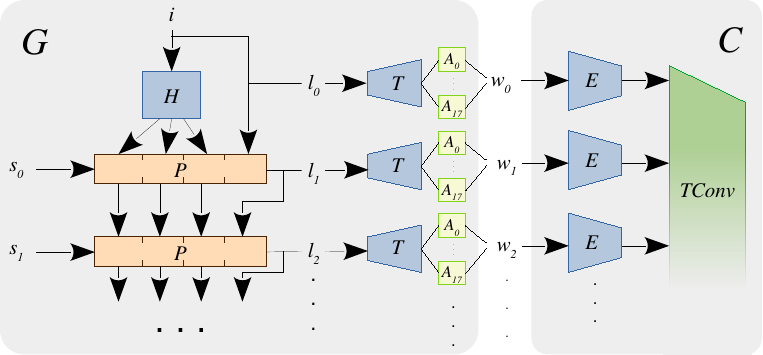}
    \caption{Our Wasserstein GAN.
     }
    \label{fig:pipeline}
\end{wrapfigure}

We train a Wasserstein GAN \cite{arjovsky2017}, consisting of generator $\generator$ and critic $\critic$.

The input to $\generator$ is a pair $(\seedinitial, \seedtemporal)$, with both $\seedinitial \sfrom \gauss^{32}$ and $\seedtemporal \sfrom \gauss^{32 \times (t - 1)}$ being Gaussian samples. The number $t$ of time steps is fixed to $25$ at training time, but can be larger at test time, because the generator is an RNN.
The output of $\generator$ is a sequence of $t$ latent codes $w_k \in\wplus{}$, with $0 \leq k < t$.
To train the generator, we first use \psp{} \cite{richardson2021}, an encoder-based inversion method for StyleGAN, to embed the training video in $\wplus{}$ space. This embedding provides the source of \enquote{real} samples for the critic to distinguish from the generator's output. No frames are rendered during training, StyleGAN is absent. This leads to considerable savings in training time and memory consumption, in particular compared to the method by Tian \etal (see \cref{sec:results}).
 Only at test time do we forward the output of our generator into StyleGAN.

Note that although we focus on demonstrating our pipeline on portrait videos,
no part of it other than the preprocessing step is inherently face-specific.

\paragraph {Data Preprocessing}
\begin{wrapfigure}{R}{0.5\textwidth}
    \includegraphics[width=0.5\textwidth]{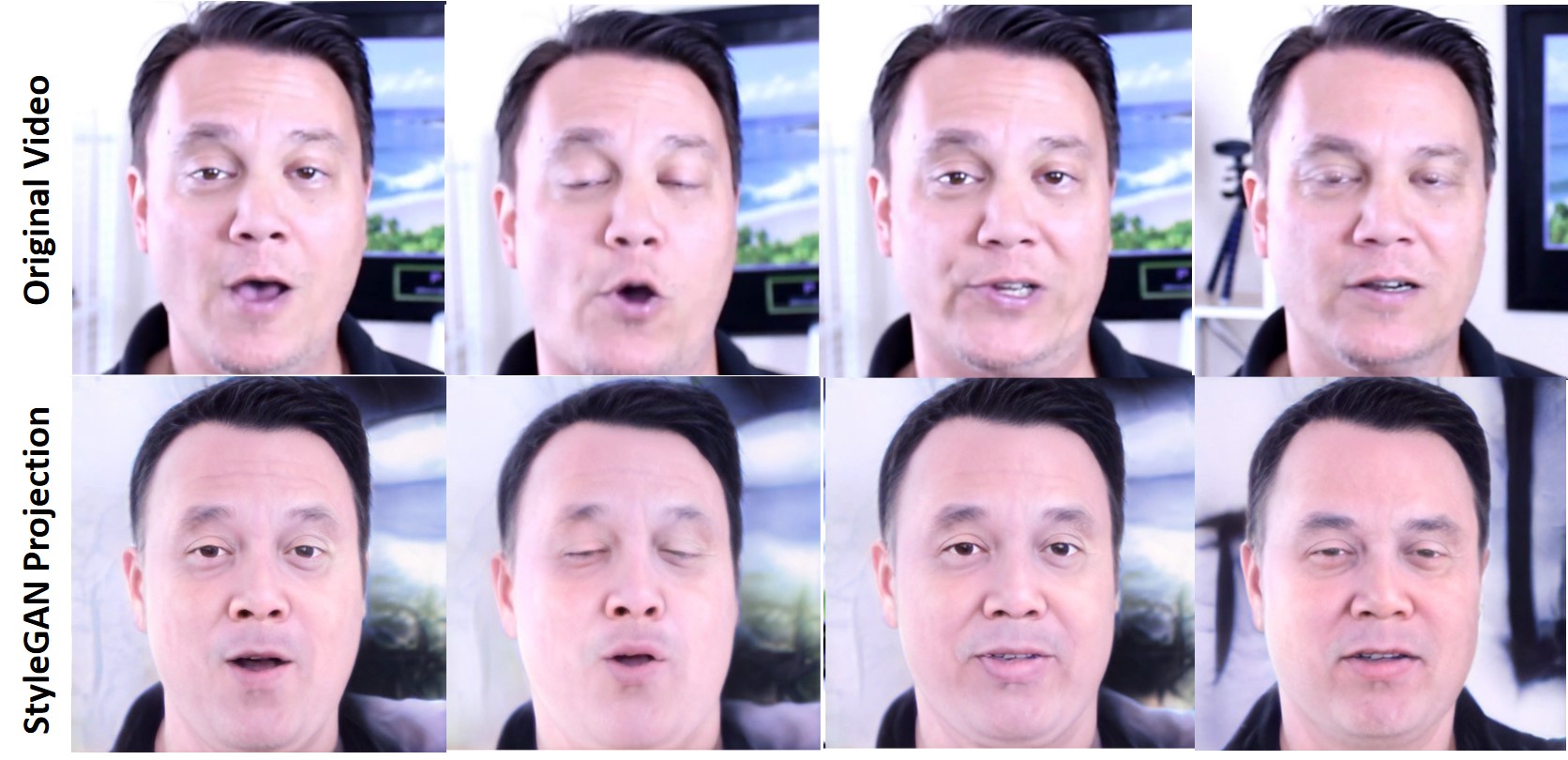}
    \caption{ Projecting face videos into $\wplus{}$ via \psp{} reasonably maintains the identity of the actor and leads to temporally smooth results.}
    \label{fig:dataset}
\end{wrapfigure}
Our training set consists of 1 single video of $< 10$ minutes, sometimes  much less. Before we can embed its frames in $\wplus{}$ via \psp{}, we preprocess them in the same way that the training data for the respective StyleGAN model has been preprocessed. In the case of faces this means that we compute face crops similarly as for FFHQ \cite{Karras2019}: Since the preprocessing there was not designed with temporal smoothness in mind, we had to slightly alter the choice of landmarks used for face alignment (we use only the eye corners, never the mouth) and applied temporal lowpass filters to the rotation and scale of the face bounding boxes. This gave decent results, best seen in our supplemental video, or in \cref{fig:dataset}.

Applying \psp{} to each frame of the training video reliably led to sequences of $\wplus{}$ codes that, when rendered with StyleGAN, showed a decently smooth video again (\cref{fig:dataset}). The identity of the training subject was not always preserved perfectly, but this is not of interest, since our goal is anyways to generate motion for a great multitude of random actors.

\paragraph{Generator}

Our generator is a stack $\producer$ of 4 GRU cells that processs the \enquote{per-time-step randomness} $\seedtemporal$. To initialize the GRU memory, we have the MLP $\hallucinator$ \enquote{hallucinate} some memory contents for the first three cells, whereas the last is initialized with $\seedinitial$:
\[
(h_{0, 0} , h_{0, 1} ,h_{0, 2} ) := \hallucinator(\seedinitial) \hspace{1cm}
h_{0, 3} := \seedinitial
\]
After this initialization, $\producer$ can produce a sequence of low-dimensional latent codes $\latent_k \in \reals^{32}$ according to the following recurrence:
\[((h_{k+1, 0}, \ldots , h_{k+1, 3}), \latent_{k + 1}) :=      \producer(s_{k} , (h_{k, 0}, \ldots , h_{k, 3}))\]
for $0 \leq k < t - 1$. Another MLP, $\translator : \reals^{32} \rightarrow \reals^{512}$ \enquote{translates} these latent codes from the space in which motion is generated to a higher dimensional one (this is reminiscent of StyleGAN's mapping network, translating from $\zspace$ to $\wspace{}$).
A set of learned affine transformations then maps to $\wplus{}$ $\subseteq \reals^{18 \times 512}$, which gives the final output $w_0 , \ldots , w_{t - 1} \in \wplus$ of our generator.

We do not claim particular novelty of this generator design. Related works \cite{tian2021,Tulyakov18,munoz2020} have presented similar designs. The novelty of our work lies in the ideas outlined in \cref{sec:intro}, i.e., supervision in $\wplus$ instead of image space, the \enquote{offset trick} , the loss functions, and the resulting massive reduction in the amount of  data and resources required for training.

Only at test time, not at training time, do we forward $w_0 , \ldots , w_{t - 1}$ to StyleGAN for rendering of actual video frames. Note that that $t$ might be considerably larger than the 25 time steps used at training time. In \cref{sec:results} we report numbers for $t = 250$ .

\paragraph{Critic}

In contrast to $\generator$, our critic  $\critic$ \cite{arjovsky2017} can rely on a fixed $t$ and thus does not need to be recurrent. Instead, we first have another 6-layer MLP $\extractor: \wplus \rightarrow \reals^{32}$ \enquote{extract} a learned set of relevant features from each $\wplus$ step and then let a temporally convolutional network consume the resulting sequence.
We derived the architecture of this network from DCGAN \cite{Radford16}, turning spatial convolutions into temporal ones and eliminating batch normalization layers, to enforce the Lipschitz constraint required for Wasserstein GANs by a gradient penalty \cite{gulrajani2017}.

\paragraph{Loss Terms \& Training}

By training we minimize
$
    \loss = \loss_\wgan + \lambda_\gp \loss_\gp +  \lambda_\gap \loss_\gap
$,
where $\loss_\wgan + \lambda_\gp \loss_\gp$ is the usual WGAN-GP loss \cite{gulrajani2017} (with $\lambda_\gp = 50$) and $\loss_\gap$ is a novel \emph{gradient angle penalty} (with $\lambda_\gap = 100$):
When training our model only with the WGAN-GP loss we have observed (\cref{sec:results}) that synthesizing videos for $t >> 25$ can lead to motion that seems to be \enquote{looping}, i.e. the same motion pattern is repeated over and over. Based on observations reported in previous work \cite{tian2021} we suspect that $\producer$ learns to simply ignore the \enquote{per-time-step randomness} $s$ and rely exclusively on $(h_{0, 0}, \ldots , h_{0, 3})$, without modifying it much in the course of the sequence. There seems to be a tendency to make $\seedinitial$ determine the entire course of the sequence, which makes looping very likely. To counteract this, we present a new loss formulation that makes sure that the gradient of the producer output with respect to $\seedtemporal$ is at least a certain fraction of the gradient with respect to $\seedinitial$. We set:
\[\loss_\gap := \left({\max \left(0, \frac{\pi}{4} -  \varphi  \right)} \right)^2; \hspace{1cm} \text{with} \hspace{1cm} \varphi := \arctan \left(  \frac{\big|\big|[\frac{\partial d}{\partial \seedtemporal_0}, \dots, \frac{\partial d}{\partial \seedtemporal_{t - 2}}]\big|\big|}{||\frac{\partial d}{\partial \seedinitial}||}   \right ) \,, \]
where $d := \norm (\latent_{t - 1} - \latent_0)$ is the normalized Euclidean distance between the last time step and the first time step generated by $\producer$. The function $\operatorname{norm}$ here normalizes the components of the difference vector according to running statistics that are tracked during training, such that we can expect the distribution of each component to have mean 0 and variance 1. The angle $\varphi$ will be close to $0°$ if the output of $\producer$ depends mostly on $\seedinitial$, which we want to prevent.
Unless stated otherwise, we trained our models with Adam \cite{diederik2015} for 350 epochs. We exponentially average the weights of the generator throughout training using a momentum of $0.995$.

\begin{figure}
    \centering
    \includegraphics[width=0.45\textwidth]{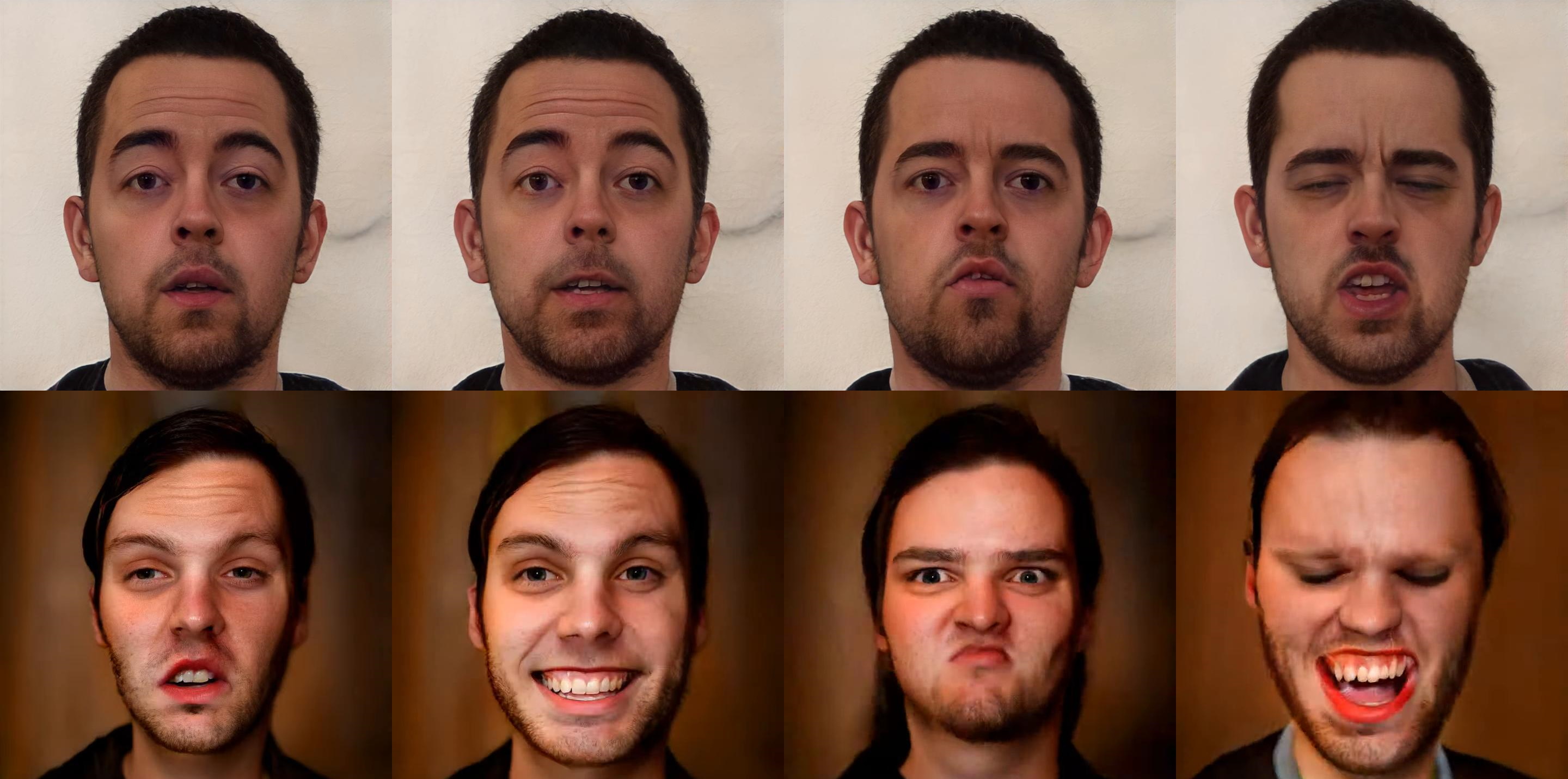} \hspace{0.5cm} \includegraphics[width=0.45\textwidth]{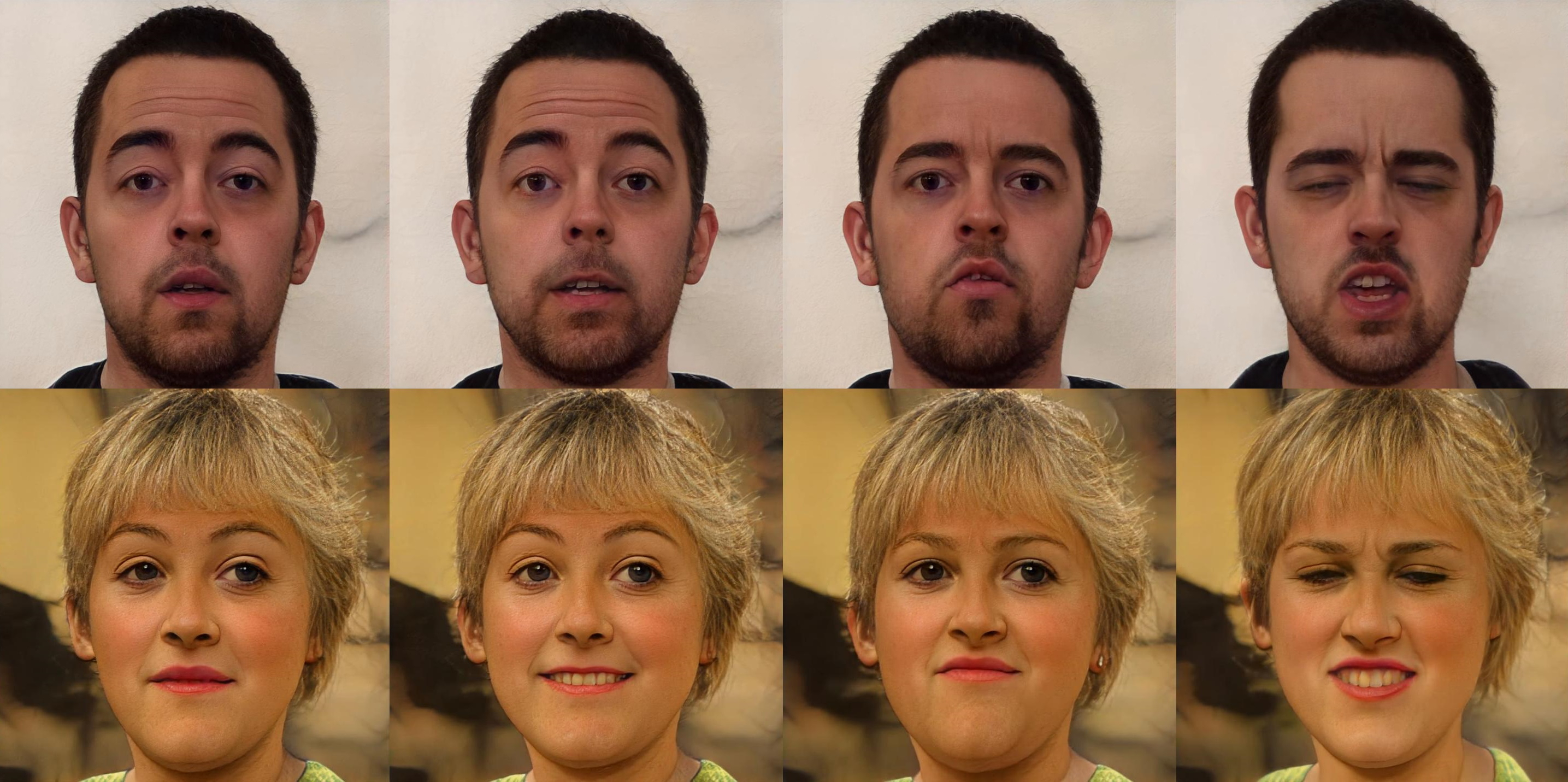}
    \caption{Top (left \& right): A motion trajectory for the training actor. Bottom left: Naively shifting the motion trajectory from the training actor to some random new actor in $\wplus$ leads to strong visual artifacts. Bottom right: Projecting the code $w_\text{new}$ of the new actor onto the PCA basis of the training actor first (\cref{sec:offset}) establishes an anchor point $w_\text{new}'$ in the training actor's point cloud. The offset by which we shift the training motion maps this anchor point to the $\wplus$ code of the new actor, reliably avoiding artifacts.
    }
    \label{fig:offset}
\end{figure}

\paragraph{The offset trick}
\label{sec:offset}
Although our network is trained on only 1 single actor, it should be able to generate motion for a large set of randomly generated actors. We achieve this by making use of the advantageous properties of StyleGAN's $\wplus$ space, that have been used for face editing before \cite{Tewari20Stylerig,shen2020,harkonen2020}: Our main assumption is that given a point in $\wplus$, the directions into which one would need to shift this point in order to change the identity of the actor that it depicts are mostly orthogonal to those directions that would change the pose/expression/articulation of the actor. If this assumption is justified (which we demonstrate in \cref{sec:results} and in our supplemental video), it should be possible to first generate a motion trajectory for our training subject and then shift this trajectory along a direction that is orthogonal to those latter directions, to transfer it to a different actor that also exists in $\wplus$.
To find the directions responsible for pose/expression/articulation we consider the $\wplus$ embedding of our training set. A simple PCA tells us those 32 directions in which the points corresponding to our training video frames extend the furthest. Since our training frames span the relevant range of motion states but always show the same actor, we can assume that shifting points in these directions changes the state, but not the identity of the actor.
Given this PCA basis and having sampled a motion trajectory $w_0 , \ldots , w_{t - 1} \in \wplus$ for our training actor, we now randomly sample a point from StyleGAN's $\zspace$ space, render it using StyleGAN and then embed it in $\wplus$ using \psp{}, obtaining $w_\text{new}$. This new point shows a random, new actor, that already is in a particular (likely nonneutral) state. For example, in the case of faces, $w_\text{new}$ might correspond to a person with the mouth closed. We must not naively use this point as the starting point for our \enquote{transferred} motion trajectory, because the motion that we generated for the training actor might start with a mouth-closing motion. Applying this motion to a mouth that is already closed would likely lead to strong artifacts (see \cref{fig:offset}). Instead, we project $w_\text{new}$ onto the PCA basis, resulting in $w_\text{new}'$. The point $w_\text{new}'$ shows our training actor in the same state as the new actor. The difference $\Delta := w_\text{new} - w_\text{new}'$ is the exact offset by which we want to shift our motion trajectory, i.e. the new trajectory is
$ w_0 + \Delta,w_1 + \Delta, \ldots , w_{t - 1} + \Delta$. For a graphical explanation of this process, please refer to our supplemental video.

As illustrated in \cref{fig:offset} and as demonstrated in our supplemental video, thanks to the disentangled representation of images in $\wplus$, this simple offset operation is sufficient to transfer motion sampled for our training actor to new random actors. Also in our supplemental video we demonstrate that not embedding the new actor with \psp{} or naively offsetting the sequence without using the PCA basis leads to much stronger artifacts.%

\section{Results}
\label{sec:results}

\begin{table*}[t]
\centering

\begin{tabular}{|c|c|c|c|c|c|}
\hline
\multirow{2}{*}{Model}                        & \multirow{2}{*}{Reference} & \multicolumn{2}{c}{FID ($\downarrow$)} & \multicolumn{2}{c|}{FVD ($\downarrow$)} \\
                                              &                                & Short      & Long       & Short      & Long       \\
\hline

\multirow{2}{*}{Ours}                            & Original     & 54.1 \plusminus 0.1 & \textbf{54.2 \plusminus 1.2} & 627.6 \plusminus 25.5 & \textbf{629.1 \plusminus 24.7}   \\
                                                 &  $\wplus{}$  & \textbf{1.1 \plusminus 0.1} & \textbf{3.9 \plusminus 1.5} & 42.9 \plusminus 12.9 & \textbf{84.0 \plusminus 17.7}      \\
\hline
\multirow{2}{*}{Ours  $\setminus $ $\loss_\gap$} & Original     & \textbf{53.9 \plusminus 0.5} & 58.8 \plusminus 4.8 & \textbf{603.7 \plusminus 39.4} & 727.1 \plusminus 159.1   \\
                                                 &  $\wplus{}$  & \textbf{1.1 \plusminus 0.0} & 7.0 \plusminus 4.2 & \textbf{33.2 \plusminus 3.7} & 178.4 \plusminus 94.4    \\
\hline
Tian   \cite{tian2021}                           & $\wspace$   & 4.07         & 97.9             & 706.3            & 2130.3           \\
Tulyakov    \cite{Tulyakov18}                    & Original     & 87.9     & 87.6      & 2849.3   & 2845.1    \\
Saito   \cite{Saito20}                           & Original     & 108.8     & 169.0     & 1211.4    & 2339.2    \\
Munoz     \cite{munoz2020}                       & Original     & 75.5     & -          & 755.4    & -          \\
\hline
\end{tabular}%
\vspace{0.2cm}
\caption{We compare FID and FVD scores between the training sets (or in the case of our method: the $\wplus{}$ embedding of the training video) and the sets of generated samples.  For a description of \enquote{Short} and \enquote{Long} see \cref{sec:metrics} . For our method we report averages and standard deviations for 5 independently trained models.
}
\label{tab:distances}
\end{table*}

\paragraph{Training data \& Metrics}
\label{sec:metrics}

For our ablation study and comparison to related work, we used sequences of faces talking into a commodity RGB camera, that were all less than 10 minutes long.
In the quantitative evaluation of trained models, we use
Fr{\'e}chet Inception Distance (FID)~\cite{Heusel17} for spatial quality, and Fr{\'e}chet Video Distance (FVD)~\cite{Unterthiner19} for the quality of motion. The reference sets for all methods are their training datasets, preprocessed as required by the particular method.  For our method we report scores in relation to both the actual original footage, as well as its $\wplus{}$ embedding (which is what our method is  trained on).

Each method was trained on the training video, depicting only one actor, as our task demands. We then generated two sets of videos from each model:
The \enquote{Short} set consists of 2048 videos that are as long as the temporal window the particular method considered at training time (see column \enquote{$t$} in \cref{fig:comparisons}).
The \enquote{Long} set consists of 128 video segments, that all have at least 128 frames. The technique by Munoz \etal{} is not able to produce samples longer than its training window, which is why \cref{tab:distances} contains no numbers for this set.
FID scores are computed on 8,000 frames randomly sampled from the reference and generated sets.
FVD scores are computed on 2048 videos from each of the two sets, with the duration of the videos again equal to the default temporal window length of each method.

We also evaluate the temporal consistency of facial identity using a variant of the Average Content Distance (ACD)~\cite{Tulyakov18}: For each generated frame, we obtain the identity features from
a popular facial recognition library~\cite{face_recog} and compute the average L2-distances between all pairs of frames of the video.
This score is then averaged over all generated videos.
\paragraph{Video Generation}

\begin{wrapfigure}{R}{0.5\linewidth}
    \centering
     \includegraphics[width=0.5\textwidth]{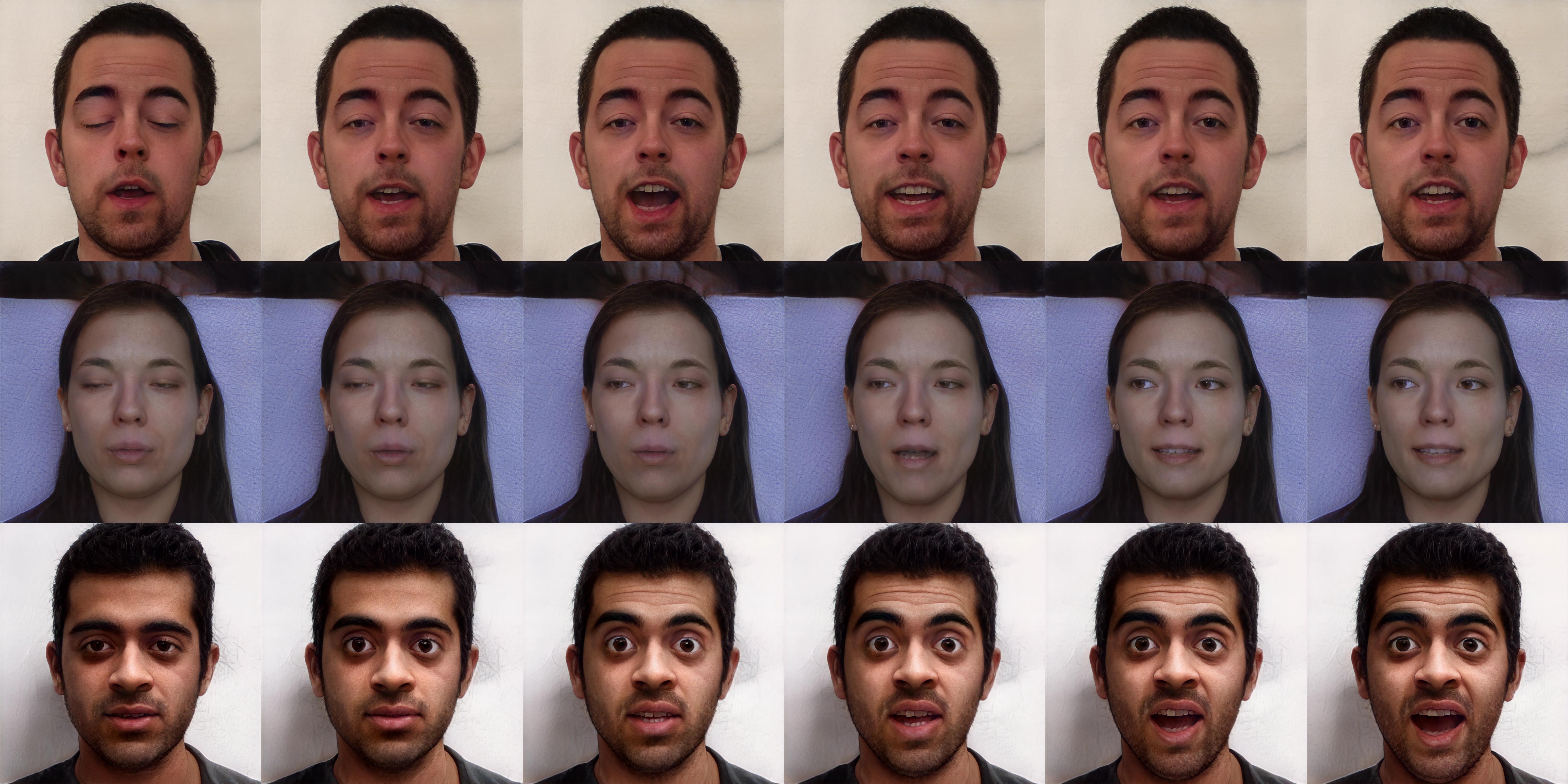}
    \caption{Videos generated by our method, for the training identities.}
    \label{fig:sameID}
\end{wrapfigure}

\cref{fig:sameID} shows sequences generated by three different models, each for the respective training identity. As shown in \cref{fig:teaser} however, the \enquote{offset trick} allows us to generate motion for randomly sampled identities as well, even though our training datasets always contain only 1 actor.
All videos are synthesized at a resolution of $1024 \times 1024$ and even though our method was trained only on a temporal window of $25$ frames, we can easily generate videos that are much longer, e.g. $1500$ frames.
The quality of motion can only be judged in our supplemental video, not on paper.
\paragraph{Comparison to Previous Methods}
\label{sec:comparisons}

\begin{table*}
\centering

\begin{tabular}{|c|>{\centering\arraybackslash} m{4.5cm}|>{\centering\arraybackslash} m{1.5cm}|c|c|}

\hline

Method & Sample & Resolution & $t$ \\
\hline

Ours (random actor)  & \includegraphics[width=4cm]{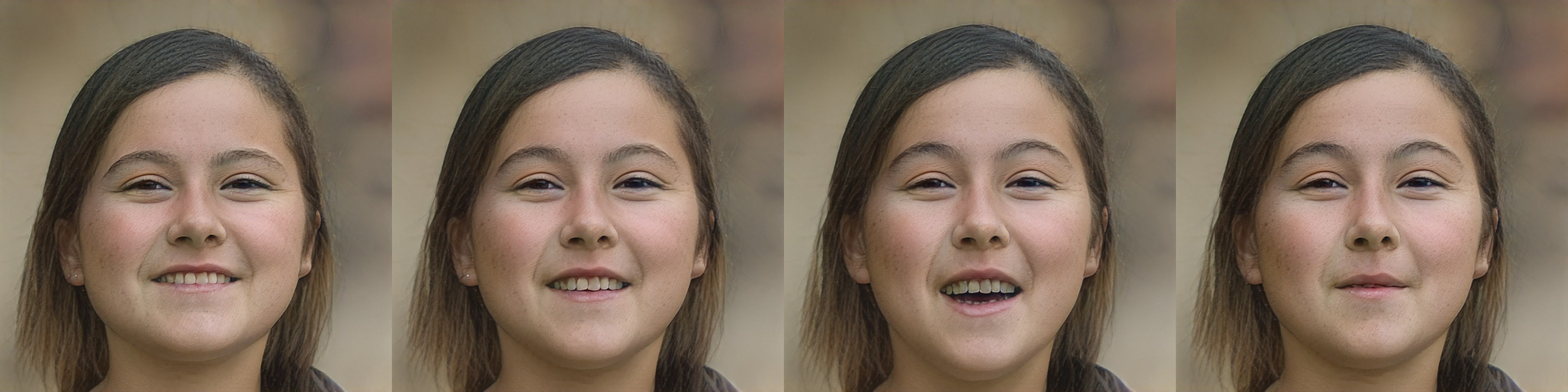} & \includegraphics[width=1cm]{res/r1024} & 25 \\
Tian \etal{} \cite{tian2021} & \includegraphics[width=4cm]{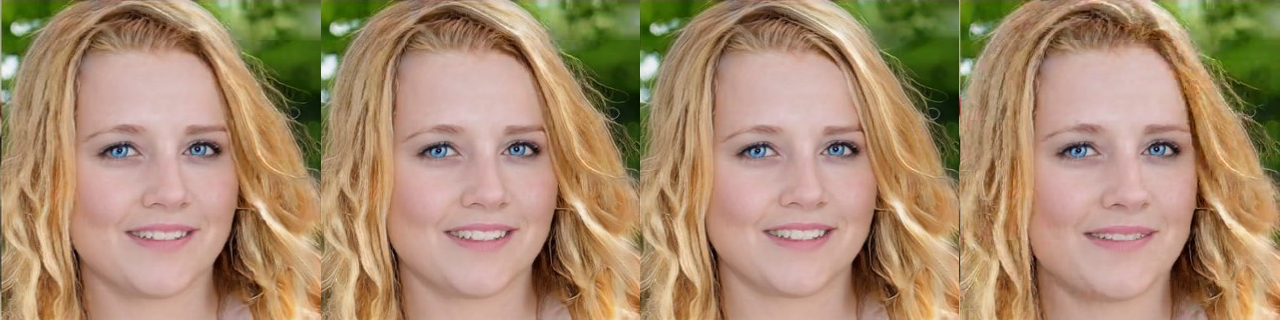} & \includegraphics[width=1cm]{res/r1024} & 16 \\
Munoz \etal{} \cite{munoz2020} & \includegraphics[width=4cm]{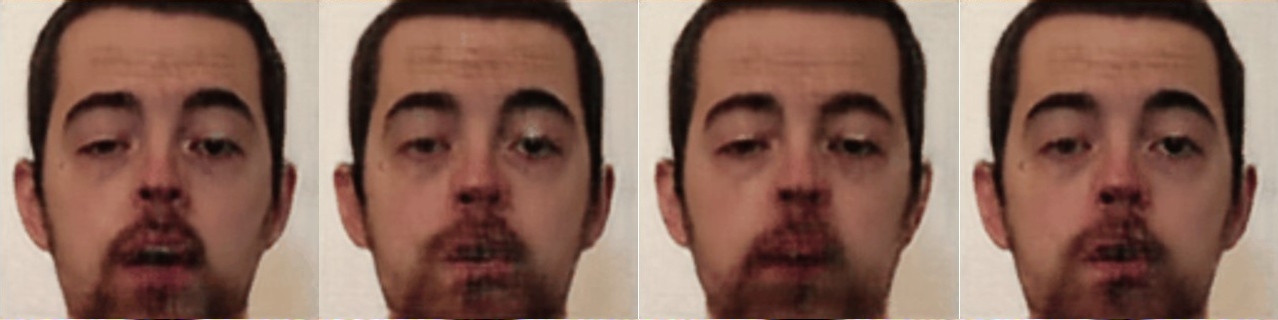} & \includegraphics[width=1cm]{res/r128} & 16 \\
Tulyakov \etal{} \cite{Tulyakov18} & \includegraphics[width=4cm]{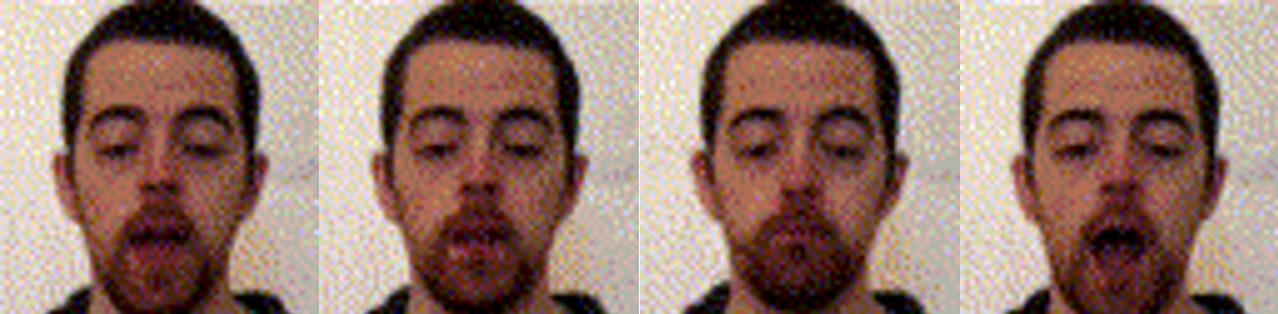} & \includegraphics[width=1cm]{res/r64} & 16 \\
Saito \etal{} \cite{Saito20} & \includegraphics[width=4cm]{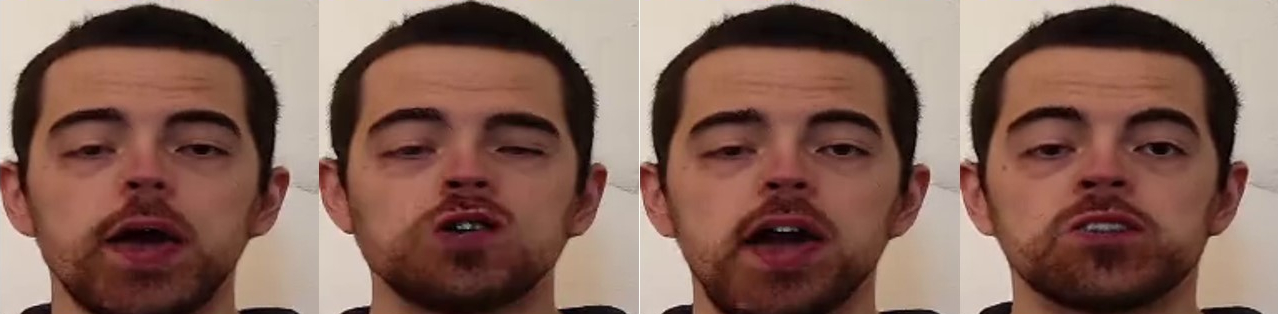} & \includegraphics[width=1cm]{res/r192} & 16 \\

\hline
\end{tabular}%
\vspace{0.2cm}

    \caption{Qualitative comparison to previous methods. Tian \etal{} does synthesize motion for the training identity by default, whereas  other previous methods can synthesize motion \emph{only} for the training identity.
    While all methods appear to generate decent spatial quality, the spatial and/or temporal resolutions of previous methods is severely limited by their computational resource demands.
    The temporal quality of the generated videos can only be judged from our supplemental video.
     }
\label{fig:comparisons}
\end{table*}

We compare to several previous techniques by training them all on the same dataset and evaluating the metrics described above.
The very recent method by Tian~\etal~\cite{tian2021} also learns a trajectory in the StyleGAN latent space, which makes it the most related to ours. We compare to their cross-domain setting, i.e. the StyleGAN generator they use was pretrained on FFHQ \cite{Karras2019}, but the motion is learned from our training video. The model that the authors kindly trained for us does by default not generate videos of the training identity, but instead samples random identities from StyleGAN's $\wspace{}$ space (\emph{not} $\wplus{}$!). This is a problem for the computation of FID and FVD scores, which always compare the generated distribution (random identities) to the training distribution (our specific training identity). We thus need to force the model to generate samples depicting our training identity, as otherwise it would have been impossible to compute fair scores.  We achieve this by sampling random frames from our training video and
projecting \cite{Karras2019} them to  $\wspace{}$. The resulting $\wspace{}$ points are then \enquote{injected} as the initial code for the model to condition its generated sequences on. As mentioned in previous works \cite{richardson2021,shen2020}, $\wspace{}$ cannot represent real images as faithfully as $\wplus{}$, which is why we did not use our training video as the reference set, but its $\wspace{}$ embedding.
Even though the method is able to produce videos at resolution $1024 \times 1024$, we show (in our supplemental material) that the model the authors trained on our dataset is not able to generate a lot of facial motion, i.e. while the camera is panning, the facial expression is very static. This is reflected in \cref{tab:distances}.
    Training this method is rather expensive and requires about $5$ days on $8$ Quadro RTX 8000 GPUs for a resolution of  $1024 \times 1024$, i.e. 40 GPU days in total.
    In contrast, our method is trained on a single Quadro RTX 8000 GPU in around 6 hours, a speedup of factor 160.
The methods by Saito~\etal~\cite{Saito20}  and  Tulyakov~\etal~\cite{Tulyakov18} generate realistic motion, but are very limited in terms of spatial resolution ($192^2$ and $64^2$ respectively). As we show in our supplemental video, the results for Munoz~\etal~\cite{munoz2020} include strong structural artifacts. All three methods are not capable of generalizing their output to random identities after being trained on only one subject, i.e. their training set would have to be much larger to make them generate the diversity of output identities Tian \etal{} and our method  achieve.
We attempted to compare to further methods \cite{Weissenborn20,Ye20}, but could not get access to their code.
Tab.~\ref{tab:distances} reports Fr\'echet distances for all methods. We have also computed ACD scores for our method (random actors) and for Tian~\etal{}: While our method averages at $5.68$ (over 5 models) Tian \etal{} achieve a score of \textbf{0.54}. We attribute this large difference to the very limited facial motion generated by Tian~\etal{}, that of course makes it much easier to preserve the identity. For visual impressions of this observation please see our supplemental video.

\paragraph{Evaluation of the Gradient Angle Penalty}

\cref{tab:distances} shows that while removing $\loss_\gap$ from our training objective slightly improves the scores for \enquote{short} samples, it considerably increases the FVD scores for \enquote{long} samples. However, since the primary purpose of $\loss_\gap$ is to prevent looping, which is maybe not effectively captured by FVD, we have recorded a very short training sequence (one single sentence, spoken three times, 20 seconds in total) that provoked strong looping artifacts in 19 out of 20 independently trained models if $\loss_\gap$ was absent, but led to looping only in 6 out of 20 models that were trained with the loss in place. This suggests that $\loss_\gap$ is indeed making looping artifacts much less likely.

\paragraph{Proof of concept: Hands}
\label{sec:hands}

To demonstrate that our method should in principle be applicable to content categories other than talking faces, we have conducted a proof-of-concept experiment for hands: We recorded the right hand of a subject for 1 hour, performing various types of motions (like showing numerals or performing a set of gestures), resulting in a dataset of around 100k frames. The only constraint was for the hand to always turn the palm to the camera and to never leave the recording space. This dataset we successfully used for training a StyleGAN model and the corresponding \psp{} inverter, both for resolution $256\times 256$. With these models available, we were able to train our temporal model with a temporal window of $75$ time steps, on several test sequences (each about 8000 frames). Results are shown in \cref{fig:teaser} and in our supplemental video.

\paragraph{Proof of concept: Cars}
\label{sec:cars}

\begin{figure}
    \centering
     \includegraphics[width=\linewidth]{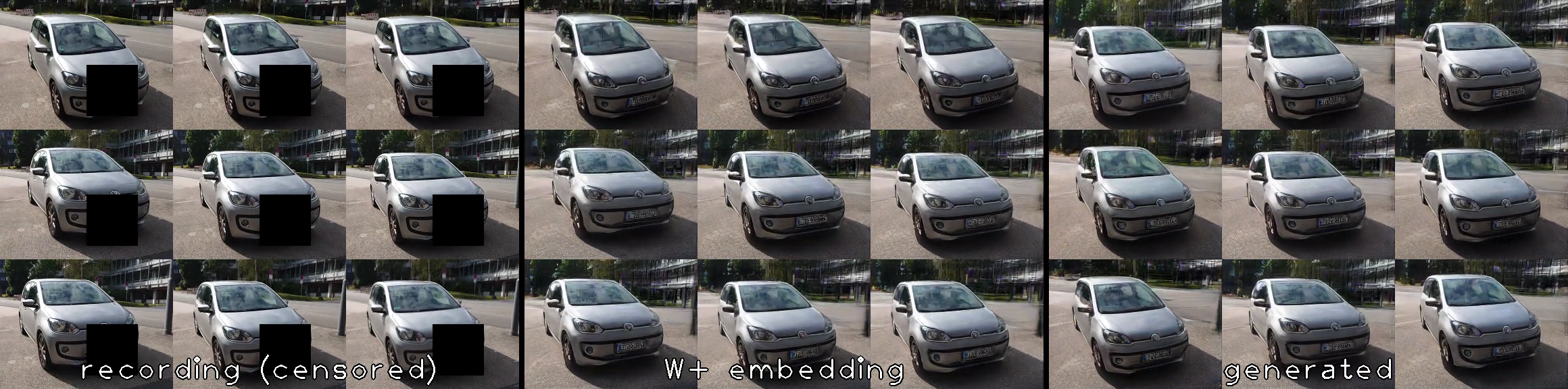}
    \caption{A synopsis of an experiment on cars: The videos we recorded (left, license plate censored) were embedded into $\wplus{}$ (center) using a \psp{} model trained on the recorded video (because we did not have a more general \psp{} of sufficient quality available). We then trained our temporal architecture on the embedding, which allowed us to generate new motion for the car (right). Results on more cars are shown in our suppplemental material.
 }
    \label{fig:cars}
\end{figure}

As a third domain, we applied our method to the category of cars. We used the official StyleGAN2 checkpoint for the LSUN-Car dataset \cite{yu2015} and trained \psp{} from scratch. LSUN-Car is a much more challenging dataset than FFHQ, because the data is not as "clean" and because there is no alignment. We thus did not manage to train \psp{} to the same level of precision as the official FFHQ checkpoints. As a result, the $\wplus{}$ embeddings contained clearly visible artifacts and the identity of the car drifted depending on the orientation. In order to nevertheless demonstrate that the core concept of our method works, we trained \psp{} a second time, but this time on the frames of our recordings. This way we easily achieved decent embeddings, showcased in \cref{fig:cars} and in our supplemental videos.
The disadvantage of this approach is that we cannot demonstrate the offset trick in this case, as our \psp{} model has only ever seen the training car and cannot embed cars randomly sampled from StyleGAN's latent space.
Of course we would have preferred to choose an object category (other than faces) for which there is a \emph{general} high-quality, temporally stable embedding method. However to the best of our knowledge, nobody has demonstrated such a method yet. We show qualitative results in \cref{fig:cars} and in our supplemental material.

\section{Conclusion}
\label{sec:conclusion}

We have presented a temporal GAN for the unconditional generation of high-quality videos.
Based on embedding the footage of only 1 actor into the latent space of StyleGAN, we are able to train our model with a minimal amount of resources and can nevertheless generate diverse motion of arbitrary length for a great number of random actors at high spatial resolution. Although these abilities also have their limitations (see supplemental), we hope that our work can pave the way for future innovations in video generation.

\paragraph{Acknowledgements}
We thank the the authors of \cite{tian2021} for training their model on the training data we sent them. We also thank Pramod Rao for his invaluable support in conducting the experiments for our evaluation section. This work was supported by the ERC Consolidator Grant 4DReply (770784).

\newpage

\bibliography{references}
\end{document}

\maketitle

\begin{figure}
\centering

\begin{tabular}{c P{0.25\linewidth} P{0.25\linewidth} P{0.25\linewidth} }
 & \idg{} & \idj{} & \ida{} \\
\raisebox{1.5cm}{Original:} & \includegraphics[width=\linewidth]{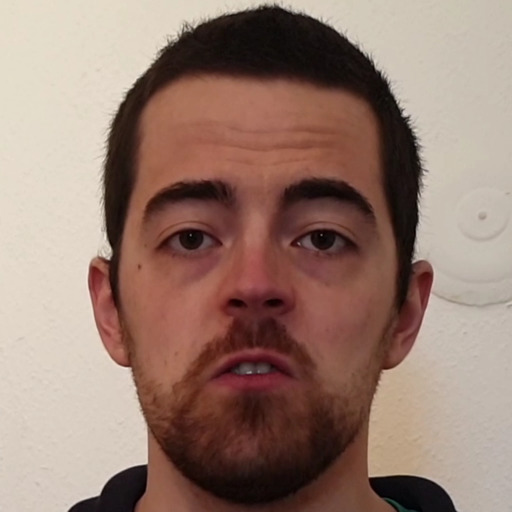} &  \includegraphics[width=\linewidth]{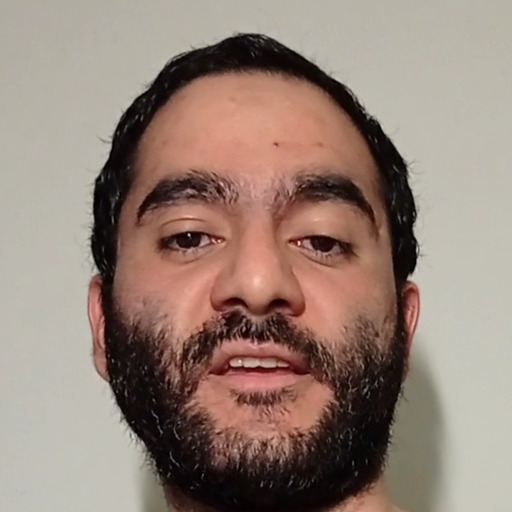} &  \includegraphics[width=\linewidth]{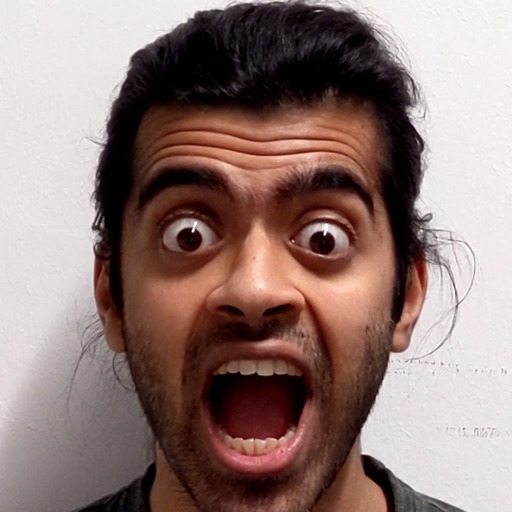} \\
\raisebox{1.5cm}{$\wplus$:} & \includegraphics[width=\linewidth]{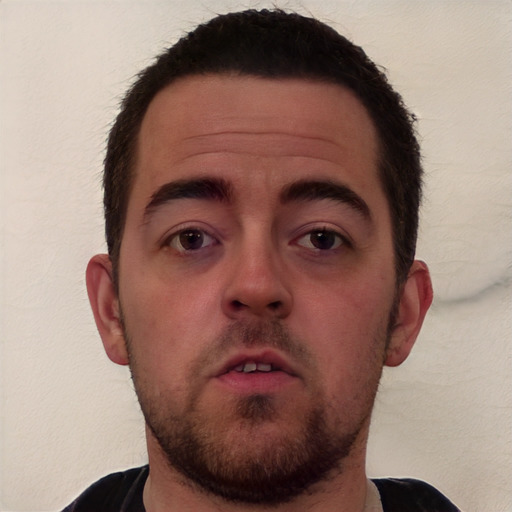} &  \includegraphics[width=\linewidth]{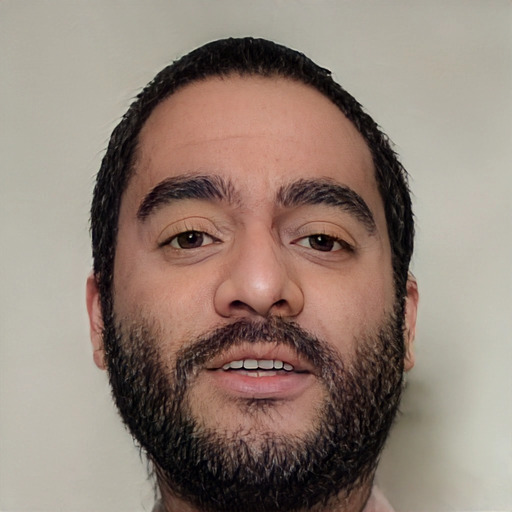} &  \includegraphics[width=\linewidth]{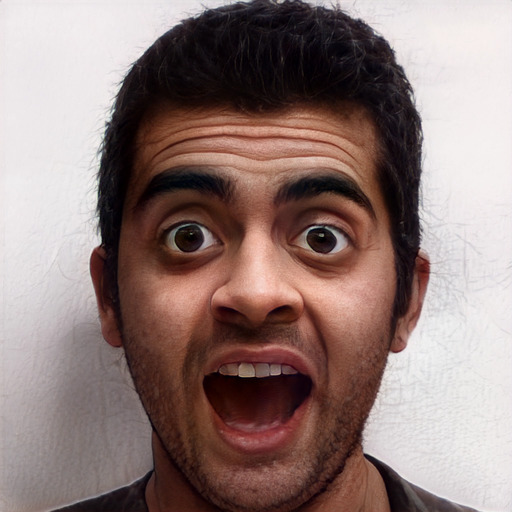} \\
Motion: & Reciting short poems. & Reading a text. & Facial expressions in random order. \\
\end{tabular}%
\vspace{0.2cm}
\caption{Screenshots from the training sequences we used for quantitative evaluation. The nubmers in this supplemental document have been computed for subjects \idj{} and \ida{}. The ones reported in the main paper are for \idg{}.}
\label{fig:identities}
\end{figure}

\section{Quantitative results for additional identities}

In \cref{tab:sj,tab:sa} we give more quantitative results, on the additional identities depicted in \cref{fig:identities}.
Our method outperforms the state of the art for these subjects as well.

\begin{table*}[h]
\centering

\begin{tabular}{|c|c|c|c|c|c|}
\hline
\multirow{2}{*}{Model}                        & \multirow{2}{*}{Reference} & \multicolumn{2}{c}{FID ($\downarrow$)} & \multicolumn{2}{c|}{FVD ($\downarrow$)} \\
                                              &                                & Short      & Long       & Short      & Long       \\
\hline

\multirow{2}{*}{Ours}                            & Original     & 62.9 \plusminus 0.2 & 65.1 \plusminus 0.8 & 846.7 \plusminus 19.1 & 944.1 \plusminus 51.7  \\
                                                 &  $\wplus{}$  & \textbf{0.6 \plusminus 0.0 } & \textbf{2.0 \plusminus 0.2} & \textbf{39.2 \plusminus 19.2} & \textbf{51.8 \plusminus 18.1 }  \\
\hline
\multirow{2}{*}{Ours  $\setminus $ $\loss_\gap$} & Original     & 62.9 \plusminus 0.1 & 64.8 \plusminus 3.0 & 848.8 \plusminus 8.0 & 926.2 \plusminus 58.5   \\
                                                 &  $\wplus{}$  & 0.7 \plusminus 0.0 & 4.3 \plusminus 2.8 & 57.0 \plusminus 93.1 & 110.7 \plusminus 74.9    \\
\hline

Tulyakov   \cite{Tulyakov18}                                                     & Original              & 76.5      & 77.8      & 1318.7    & 1338.7    \\
Saito   \cite{Saito20}                                                           & Original              & 41.5     & 51.6      & 640.0     & 935.2     \\
Munoz    \cite{munoz2020}                                                        & Original              & 46.4      & -          & 578.3     & -          \\

\hline
\end{tabular}%
\vspace{0.2cm}

\caption{FID and FVD scores for subject \idj{}. All metrics were computed as described in the main paper.}
\label{tab:sj}
\end{table*}

\begin{table*}[h]
\centering

\begin{tabular}{|c|c|c|c|c|c|}
\hline
\multirow{2}{*}{Model}                        & \multirow{2}{*}{Reference} & \multicolumn{2}{c}{FID ($\downarrow$)} & \multicolumn{2}{c|}{FVD ($\downarrow$)} \\
                                              &                                & Short      & Long       & Short      & Long       \\
\hline

\multirow{2}{*}{Ours}                            & Original     & 52.7 \plusminus 0.2 & 53.7 \plusminus 0.5 & 589.2 \plusminus 9.7 & 625.3 \plusminus 3.1  \\
                                                 &  $\wplus{}$  & 3.7 \plusminus 0.2 & \textbf{4.8 \plusminus 0.4} & 61.4 \plusminus 4.2 & \textbf{98.6 \plusminus 11.3}  \\
\hline
\multirow{2}{*}{Ours  $\setminus $ $\loss_\gap$} & Original     & 52.3 \plusminus 0.5 & 55.2 \plusminus 2.1 & 590.9 \plusminus 15.2 & 679.0 \plusminus 28.9  \\
                                                 &  $\wplus{}$  & \textbf{3.4 \plusminus 0.1} & 8.2 \plusminus 2.0 & \textbf{54.0 \plusminus 3.6} & 133.5 \plusminus 28.5  \\
\hline

Tulyakov   \cite{Tulyakov18}                                                     & Original             & 123.0     & 141.1     & 1163.3    & 1500.3     \\
Saito   \cite{Saito20}                                                           & Original             & 82.1      & 270.3     & 823.9     & 2090.8   \\
Munoz    \cite{munoz2020}                                                        & Original               & 83.3      & -          & 1037.0    & -           \\

\hline
\end{tabular}%
\vspace{0.2cm}

\caption{FID and FVD scores for subject \ida{}. This subject was not talking, but instead performing some simple face motions in a random order (like smiling or acting surprised). The training video is only 1 minute and 20 seconds in length.}

\label{tab:sa}
\end{table*}

\section{Experimental Details}
\label{sec:add_results}

\begin{table*}
\centering
\begin{tabular}{|l|l|l|l|l|}
\hline
                                 & GPU Used         & GPU Memory & Training time (max) \\ \hline
Ours                             & Quadro RTX 8000  & 48 GB (8GB used)     & approx. 6 hours     \\ \hline
Tulyakov~\etal~\cite{Tulyakov18} & GeForce RTX 2080 & 12 GB      & approx. 6 hours     \\ \hline
Munoz~\etal~\cite{munoz2020}       & Quadro RTX 8000  & 48 GB      & approx. 2 days      \\ \hline
Saito~\etal~\cite{Saito20}       & GeForce RTX 2080 & 12 GB      & approx. 15 hours    \\ \hline
\end{tabular}
\vspace{0.2cm}

\caption{Training details for the various methods. The authors of Tian~\etal~\cite{tian2021} trained their method for us. %
}
\label{tab:training}
\end{table*}

\subsection{Training details}

We trained all methods with their default hyperparameters, except for Saito et al, where we had to adjust the batch size to $2$ and set \texttt{clstm channels = 512} . The authors of Tian~\etal~\cite{tian2021} kindly trained their technique on the training data we sent them.  We trained all methods with at least the computational resources that our method uses, but usually gave them much more training time. \Cref{tab:training} gives a comprehensive overview.

Each training video contains 1 single actor/object. We computed quantitative evaluations for 3 actors, reported in the main paper and this supplemental. For the proof of concept on hands, all training data was recorded from one actor. In the case of cars, we show qualitative results of 3 different cars. We use a batch size of $128$, learning rate of $0.005$, and exponential averaging of the weights with a momentum of $0.997$ in all experiments. All temporal generative networks are trained for $350$ epochs.

\subsection{Evaluation details}

FID scores have been computed by sampling $8000$ frames from both the training set (as preprocessed for the particular method) and the set of generated videos. We used the FID implementation in \url{https://github.com/mseitzer/pytorch-fid}.

FVD scores have been computed by sampling $2048$ video slices from both the training set (as preprocessed for the particular method) and the set of generated videos. For each method, and regardless of the length of the samples we generated (\enquote{short} versus \enquote{long}), the videos we sampled were always 25 frames long for our method and 16 frames long for the previous methods. We used the FVD implementation in

\begin{center}
{\small\url{https://github.com/google-research/google-research/tree/master/frechet_video_distance}}
\end{center}

ACD scores have been computed always on 128 \enquote{long} samples drawn from the trained models. For our method these long samples were 400 frames long. For Tian \etal{} they were 128 frames long. Having a good ACD becomes harder as sequences grow longer.

\section{Limitations}
\label{sec:limitations}

Even though we are able able to further to the state of the art in video generation, in particular with respect to the amounts of computational resources and training data necessary to generate a large amount of diverse videos, our method has several limitations:
For one, the quality of our generated videos strictly depends on the quality of the underlying StyleGAN model and its corresponding \psp{} inverter. For example, in the case of faces we have observed that nontrivial video backgrounds tend to not be represented in a temporally stable way. The importance of temporally stable embedding is also underlined by an experiment we made with a BigGAN model \cite{Brock19} instead of a StyleGAN model: We used a SOTA optimization-based method \cite{huh2020} to embed several short videos (to the best of our knowledge there are no encoder-based inversion methods, see \cite{xia2021}). The embeddings contain strong temporal noise, making training our method pointless (see supplemental video). For the sources of the videos, see \cref{tab:sources}

Another limitation is the fact that while our approach does not contain any inherently face-specific components and even though we are showing a proof-of-concept for animating hands and cars, it is still unclear whether all the advantageous properties of StyleGAN's $\wplus$ space can be made use of in any arbitrary domain, e.g. if our offset trick will work will work there.

\section{Detailed architecture}

For the sake of completeness, \cref{tab:arch:initer,tab:arch:gru,tab:arch:translator,tab:arch:critic} give detailed specifications for the architecture components that we outlined in Figure 2 of the main paper.

\begin{table*}[h]
\centering
{
\begin{tabular}{clc}
Input                          & Module            & Outputs (Dimensionality) \\ \hline
$\seedinitial $ & 4 layers (Fully Connected + LeakyReLU)                    & $m$   ($3 \times 32$)      \\
$m$  & BatchNorm            & $(h_{0, 0} , h_{0, 1} ,h_{0, 2} ) $ ($3 \times 32$)         \\
\end{tabular}}
\caption{The \enquote{hallucinator} $\hallucinator$ is responsible for producing some initial contents for the GRU memory. For each one of the four stacked GRU cells it produces a vector of length $32$.}
\label{tab:arch:initer}
\end{table*}
\begin{table*}[h]
\centering
{
\begin{tabular}{clc}
Input                          & Module            & Outputs  (Dimensionality) \\ \hline
$s_{0} , (h_{0, 0}, \ldots , h_{0, 3})$ & GRU (4 stacked cells)                      & $(h_{1, 0}, \ldots , h_{1, 3}), \latent_{1}$    ($3 \times 32, 32$)   \\
$s_{1} , (h_{1, 0}, \ldots , h_{1, 3})$ & GRU (4 stacked cells)                      & $(h_{2, 0}, \ldots , h_{2, 3}), \latent_{2}$   ($3 \times 32, 32$)    \\
$s_{2} , (h_{2, 0}, \ldots , h_{2, 3})$ & GRU (4 stacked cells)                      & $(h_{3, 0}, \ldots , h_{3, 3}), \latent_{3}$    ($3 \times 32, 32$)   \\

$\dots$ & $\dots$ & $\dots$
\end{tabular}}
\caption{The feature generator $\producer$ consists of four stacked GRU cells. Its hidden state is initialized with the output of $\hallucinator$ (\cref{tab:arch:initer}) and it translates a sequence of random vectors $\seedtemporal_k$ into intermediate latent codes $\latent_{k+1}$ for $0 \leq k < t - 1 $ in a recurrent fashion.}
\label{tab:arch:gru}
\end{table*}
\begin{table*}[h]
\centering
{
\begin{tabular}{cp{0.45\linewidth}c}
Input                          & Module            & Outputs (Dimensionality) \\ \hline
$\latent_k$ & BatchNorm + Affine transform + PixelNorm                & $\latent'_k$ (512)       \\

$\latent'_k$ & 4 layers (FullyConnected + LeakyReLU)                   & $v'_k$ (512)       \\
$v'_k$ & BatchNorm + Affine transform                & $v_k$(512)       \\

$v_k$ & 18 parallel layers (FullyConnected + LeakyReLU + BatchNorm)       & $w_k$ (18 $\times$ 512)
\end{tabular}}
\vspace{0.2cm}

\caption{The latent mapper $\translator$ is an MLP that transforms the outputs $\latent_k$ of the feature generator $\producer$ into  StyleGAN latent codes $w_k \in \wplus$: After a 4-layer MLP that widens the dimensionality from 32 to 512, we employ 18 independent fully connected layers (with LeakyReLu activation), in a way similar to how StyleGAN broadcasts its $\wspace$ vectors to its 18 Style layers.
}
\label{tab:arch:translator}
\end{table*}
\begin{table*}[h!]
\centering
{
\begin{tabular}{cp{0.45\linewidth}c}
Input                          & Module & Outputs (Dimensionality) \\ \hline
$w_k$ & FullyConnected + LeakyReLU (2 layers)               & $e'_k$ (512)          \\
$e'_k$ & FullyConnected + LeakyReLU (4 layers)       & $e_k$ (32)            \\

$e_0, \ldots, e_{t - 1}$ & 1D-version of the DCGAN critic \cite{Radford16}, with 32 input channels  & Critic Output (1)
\end{tabular}}
\vspace{0.2cm}

\caption{The latent critic takes a sequence $w_0 , \ldots , w_{t - 1}$ of StyleGAN latent codes as input, and produces a scalar output.
}
\label{tab:arch:critic}
\end{table*}

\clearpage

\begin{table*}
\centering
\begin{tabular}{cl}
BigGAN category & YouTube key \\
\hline
\texttt{indigo bird (014)} & \texttt{DZOiCmxSU2k} \\
\texttt{green mamba (064)} & \texttt{hG4Wvp0U18A} \\
\texttt{bison (347)} & \texttt{L4eOhuLDfeU} \\
\texttt{gazelle (353)} & \texttt{jMIiB9DnRXg} \\
\end{tabular}
\vspace{0.2cm}
\caption{The sequences we embedded into the BigGAN latent space were excerpts of YouTube videos.}
\label{tab:sources}
\end{table*}

\bibliography{references}